\documentclass{ecai} 

\usepackage{latexsym}
\usepackage{amssymb}
\usepackage{amsmath}
\usepackage{amsthm}
\usepackage{booktabs}
\usepackage{enumitem}
\usepackage{graphicx}
\usepackage{color}
\usepackage{flushend}

\newtheorem{remark}{Remark}

\newcommand{\BibTeX}{B\kern-.05em{\sc i\kern-.025em b}\kern-.08em\TeX}

\begin{document}


\begin{frontmatter}


\paperid{123} 


\title{Recovering implicit physics model under real-world constraints}


\author[A]{\fnms{Ayan}~\snm{Banerjee}\orcid{0000-0001-6529-1644}}
\author[B]{\fnms{Sandeep K.S.}~\snm{Gupta}\orcid{0000-0002-6108-5584}\thanks{Email: abanerj3@asu.edu, sandeep.gupta@asu.edu. Funded in part by DARPA AMP-N6600120C4020, and FIRE - P000050426 and NSF FDT-BioTech 2436801}.}
\address[A,B]{IMPACT Lab, Arizona State University}


\begin{abstract}
Recovering a physics-driven model, i.e.\ a governing set of equations of the underlying dynamical systems, from the real-world data has been of recent interest. Most existing methods either operate on simulation data with unrealistically high sampling rates or require explicit measurements of all system variables, which is not amenable in real-world deployments. Moreover, they assume the timestamps of external perturbations to the physical system are known a priori, without uncertainty, implicitly discounting any sensor time-synchronization or human reporting errors. In this paper, we propose a novel liquid time constant neural network (LTC-NN) based architecture to recover underlying model of physical dynamics from real-world data. The automatic differentiation property of LTC-NN nodes overcomes problems associated with low sampling rates, the input dependent time constant in the forward pass of the hidden layer of LTC-NN nodes creates a massive search space of implicit physical dynamics, the physics model solver based data reconstruction loss guides the search for the correct set of implicit dynamics, and the use of the dropout regularization in the dense layer ensures extraction of the sparsest model. Further, to account for the perturbation timing error, we utilize dense layer nodes to search through input shifts that results in the lowest reconstruction loss. Experiments on \textbf{four} benchmark dynamical systems, three with simulation data and one with the real-world data show that the LTC-NN architecture is more accurate in recovering implicit physics model coefficients than the state-of-the-art sparse model recovery approaches. We also introduce \textbf{four} additional case studies (\textbf{total eight}) on real-life medical examples in simulation and with real-world clinical data to show effectiveness of our approach in recovering underlying model in practice. 
\end{abstract}

\end{frontmatter}


\vspace{-0.1 in}\section{Introduction}
Model recovery problem concerns with deriving underlying physics driven governing equations of a system from data~\cite{quade2018sparse}. Different from model learning, the model recovery has two goals: a) to accurately reconstruct the data, and b) derive the fewest terms required to represent the underlying nonlinear dynamics. As such sparse identification of nonlinear dynamics is needed~\cite{kaiser2018sparse,kaheman2020sindy}. It is useful for several applications such as learning digital twins~\cite{WANG2023TWIN}, safety analysis~\cite{banerjee2023statistical}, anomaly detection~\cite{maity2023detection}, explainable artificial intelligence (AI)~\cite{heaton2023explainable} and prediction~\cite{regis2023data}. Two broad categories of techniques exist (Table \ref{tbl:relworks}): a) using physics driven deep learning on large datasets, and b) using transformations that linearize the nonlinear dynamics using the Koopman theory~\cite{koopman2023ul} and then performing linear regression with sparsity constraints. It is generally acknowledged that both categories of techniques suffer significant performance degradation on data from real-world systems~\cite{o2023investigating}. Some effort has been undergoing to improve performance of both categories of approaches on systems with limited data and noise~\cite{chen2021physics,fasel2022ensemble}, however, such problems are only a small subset of issues observed in real-world deployments.  

This paper focuses on the model recovery problem that arises in real-world systems which may include interactions with the human users at runtime. As such we consider a control affine system whose $n$ dimensional state space $X = \{x_1\ldots x_n\} \in \mathcal{R}^n$ is given by:
\begin{equation}
\label{eqn:sys}
\scriptsize
\dot{X} = f(X,\Theta) + g(X,\Theta) (U + U_{ex}),
\end{equation}
\noindent where $f(X,\Theta): \mathcal{R}^n \times \mathcal{R}^p \to \mathcal{R}^n$ is a model of the natural unperturbed dynamics of the physical system with human users in it that is perturbed by: a) a autonomous system $U = \mathcal{K}(X)$, where $\mathcal{K}: \mathcal{R}^n \to \mathcal{R}^m$ is a control function that generates the $m$ dimensional actuation signals to the physical system and b) input from the human user $U_{ex}\in \mathcal{R}^m$. We denote the total perturbation as $U_T = U + U_{ex}$. $g(X,\Theta): \mathcal{R}^n \times \mathcal{R}^p \to \mathcal{R}^n\times\mathcal{R}^m $, expresses the effect of the input perturbation. $\Theta$ is the set of model coefficients of the dynamical system. 

Real-world observations are limited by additional constraints ($Ci$). 

\noindent{\bf C1: Low sampling rate -} Storage and energy constraints in real-world deployments often limit sensing frequency~\cite{Banerjee12Ensuring}. As 
 per the Nyquist Shannon sampling theory~\cite{vaidyanathan2001generalizations}, full information about the model coefficients $\Theta$ is available in observation $X$ if it is sampled at the Nyquist rate i.e.,the sensing frequency is twice the maximum frequency in the power spectrum of $X$ obtained from Eqn.\@\ref{eqn:sys}. The control input $U$ has the same sampling frequency as $X$. However, in practice, the sensor data is logged at a sub-Nyquist rate. 

\noindent{\it Difficulty in model recovery with C1}: A low sampling frequency causes performance degradation of model recovery~\cite{champion2019discovery}. As samples grow further apart in time, the set of nonlinear functions that accurately fit the samples becomes larger making it increasingly difficult to find the correct underlying model in the expanded search space. 

\noindent{\bf C2: Perturbed system -} Since in deployment the system operation cannot be disrupted, the traces $X$ are always from a perturbed system following Eqn\@\ref{eqn:sys} and there is no trace obtained from the unperturbed part of the system. Plus, there can be, $q > 0$, external human inputs $U_{ex}$ at times $t_1 \ldots t_q, t_i \in \mathcal{R}^{\geq 0},\forall i$ that are sparse in time and are described by a set of tuples $Ex = \{(U^1_{ex},t_1),\ldots (U^q_{ex},t_q)\}$, 
Model recovery from perturbed systems with control inputs has been considered in the literature, however, to the best we know, sparse external human inputs have been ignored.

\noindent{\it Difficulty in model recovery with C2:} While input perturbations due to the control logic are Lipschitz continuous, the external human inputs are discontinuous that introduce transients in the dynamical system. Approaches such as least square minimization based model recovery~\cite{kaiser2018sparse} that ignores human inputs may erroneously introduce higher order nonlinear terms in the model due to such transients.

\noindent{\bf C3: Sparsity structure in high dimensional nonlinear function space -} The model in Eqn.\@\ref{eqn:sys} is a physics driven model, which is derived from first principles. As such the algebraic structure $S$ of this model is sparse in the nonlinear function space. However, in real-life systems, the sparsity information is often unknown.

\noindent{\it Difficulty in model recovery with C3:} Sparsity requirements add additional constraints to the model coefficients $\Theta$ leading to a constrained optimization problem which may be more complex than unconstrained versions~\cite{bertsekas2014constrained}. Solutions such as physics-driven deep learning use original model coefficients in the loss function (see Table \ref{tbl:relworks}), which is not available in real-world deployments.

\noindent{\bf C4: Implicit dynamics -} Privacy requirements, cost of sensing, and resource constraints on storage results in measurements of only a subset of the system variables $X$ in real-world deployments. This constraint is expressed as a sensing matrix $C$, which is a diagonal matrix, where $c_{ii} = 1$ if a sensor senses $x_i$ and $c_{ii} = 0$ if there is no sensor for $x_i$. Hence, only sampled traces of $Y = CX$ is available as sensed data and some physics-driven dynamics model is implicit. All actuation inputs $U$ and user inputs $U_{ex}$ are sensed.

\noindent{\it Difficulty in model recovery with C4:} With the implicit dynamics, the problem formulations of existing model recovery approaches fall apart. To solve this, existing techniques increase the state variable dimension by adding new variables that are differentials of the measured state variables. We refer to this as {\it weakly} implicit dynamics. There are two problems with such approaches: a) in real-world deployment differential computation can be inaccurate especially with low sampling rates, and b) to model the nonlinearity of unmeasured state variables, higher order differentials need to be computed, which not only increases the model recovery error but also significantly increases the dimensionality of the model recovery problem.

\noindent{\bf C5: Input uncertainty -} Timestamps of the input perturbations are uncertain in the real world due to several reasons: a) faulty reporting by human users, and b) the inputs and the state variables are often measured by different sensors which may have uncertain differences in clock times~\cite{o2021longitudinal}. In other words, the measured human inputs $U^s_{ex}=U_{ex}+\eta^u_{ex}$ has unknown error $\eta^u_{ex}$ and the measured user input activation time $t^s_{ex}=t_{ex}+\eta^t_{ex}$ has unknown error $\eta^t_{ex}$. 

\noindent{\it Difficulty in model recovery with C5:} Current approaches do not consider the noise in the timestamps and only focus on the noise in magnitude. Incorporating temporal uncertainties can intractably increase the search space of the model recovery problem.

Given a set $\mathcal{T}$ of traces of $Y,U,U_{ex}$ over time, in this paper we solve the model recovery problem to derive $\Theta_{est}$ such that $||\Theta - \Theta_{est}||^2 < \epsilon$, for some $\epsilon > 0$ error limit. {\it However, note that the original $\Theta$ is unknown and cannot be used in the solution of the problem}. Hence, the estimated model coefficients $\Theta_{est}$ has to be utilized to first derive an estimated trace $Y_{est}$ and then $||Y_{est}-Y||^2$ can be used as objective function.

\begin{table*}
	\centering
	\scriptsize
	\caption{Related works in model recovery. High in column 2 means greater than double the Nyquist rate, Low means at Nyquist rate.}
	\begin{tabular}{|p{1.3 in}|p{0.4 in}|p{0.5 in}|p{0.5 in}|p{0.5 in}|p{0.5 in}|p{0.5 in}|p{1.5 in}|}
	 \hline
		{Approach} & {C1: Sampled data} & { C2: Control Perturbed } & { C2: Human Perturbed } &{C3: Sparse model} & {C4: Implicit dynamics} & {C5: Input uncertainty} & {Assumptions} \\ \hline
Eigen system Realization ~\cite{oymak2021revisiting}& Low & Yes & No &	No & Yes &	Magnitude  & Linear system \\\hline
Genetic Algorithm~\cite{schmidt2009distilling} & High & No & No & Yes &	No & No & Low dimensional nonlinear systems\\\hline
SINDy~\cite{quade2018sparse} & High  & No  & No & Yes   & No & No  & Known sparsity threshold \\\hline
SINDYc~\cite{kaiser2018sparse} & High & Yes & No & Yes &	Weak & Magnitude & same as above\\\hline
E-SINDY~\cite{fasel2022ensemble} & Low & Yes &  No & Yes & Weak & Magnitude & same as above \\\hline
NODE + structure ~\cite{Kookjin21Machine} & High & No & No & Yes &	No & No & Known metriplectic structure\\\hline
PINNs + Sparse Regression~\cite{chen2021physics} & Low & No & No & Yes &	No & No & Physics loss for original coefficients\\\hline
\textbf{This paper} & \textbf{Low} & \textbf{Yes} & \textbf{Yes} & \textbf{Yes} & \textbf{Yes} & \textbf{Yes} & \textbf{Black box ODE solver in the loss}\\\hline
	\end{tabular}
	\label{tbl:relworks}
\end{table*}

\vspace{-0.1 in}\subsection{Approach and contributions}
\begin{figure}[t]
\center
\includegraphics[clip=false,trim=0 70 0 0,width=\columnwidth]{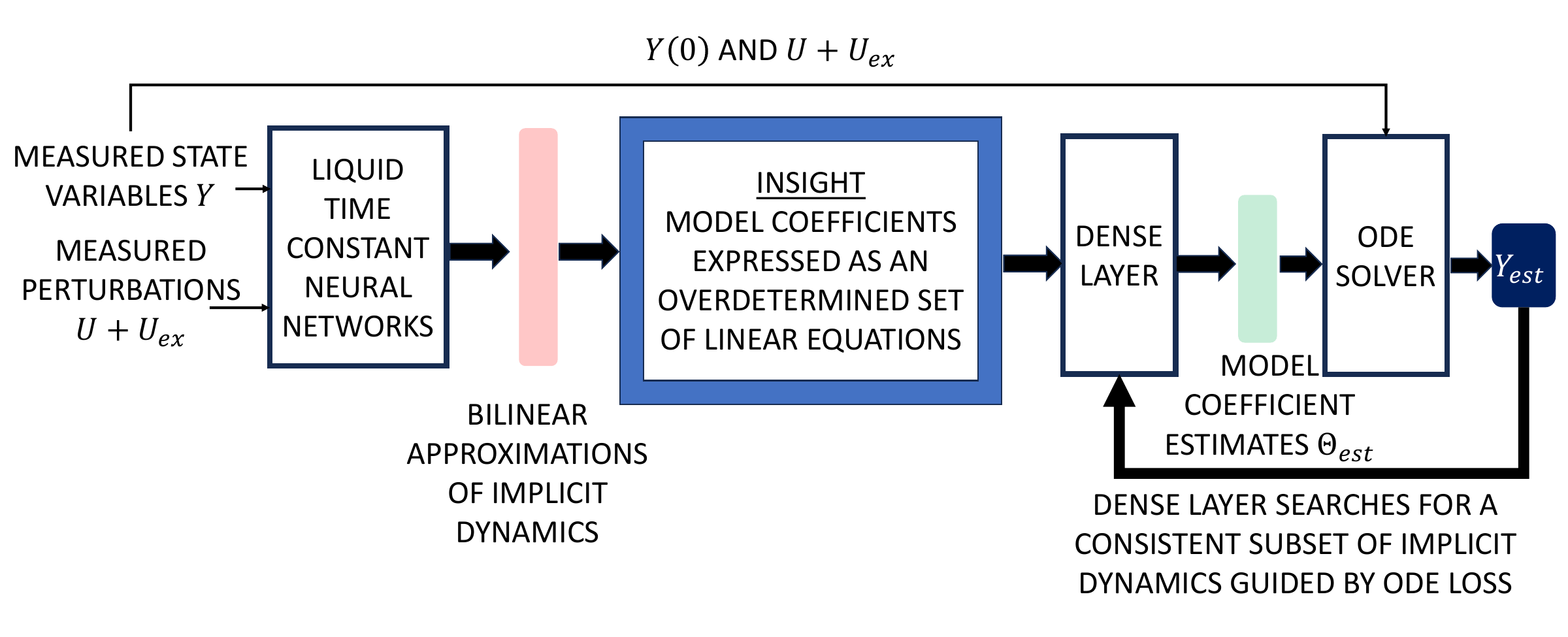} 
 \caption{Liquid Time Constant neural network based architecture for model recovery.}
    \label{fig:Approach}
\end{figure} 
Our \textbf{main contribution} is the extension of liquid time constant neural network (LTC-NN) to advanced neural structure (LTC-NN-MR) for recovering physics model under real-world constraints (Fig.\ \ref{fig:Approach}). 

\noindent{\bf Intuition for addressing C1:} The automatic differentiation property of LTC-NN class of neural architectures~\cite{banerjee24Emily}, which also includes continuous time recurrent neural networks (CT-RNN) or neural ordinary differential equations (NODE), can solve a system of ODEs with arbitrary temporal resolution. Hence, at low sampling rates, LTC-NN can accurately represent system dynamics with arbitrary precision in between samples while maintaining the structure. 

\noindent{\bf Intuition for addressing C2:} The forward pass of LTC-NN has the same form as the control affine dynamics in Eqn.\@\ref{eqn:sys}. Hence, each node inherently decouples the perturbed and un-perturbed dynamics of the system and learns them separately. 

\noindent{\bf Intuition for addressing C3 \& C4:} The forward pass of LTC-NN can take the form of bi-linear approximations of the implicit dynamics in Eqn.\@ \ref{eqn:sys} and hence it can search through the space of implicit dynamics (Fig. \ref{fig:Approach}). The measurements of $Y$, can be used to convert the set of implicit dynamics to an over-determined system of equations that are linear in terms of the model coefficients. As such an over-determined system of equation may have no solution unless either some equations are rejected or are expressed as linear superposition of other equations. To search for a set of consistent equations to estimate the model coefficients, a dense layer is utilized. The search process of the dense layer is guided by a loss function (\textit{ODEloss}) that computes the mean square error between the reconstructed (estimated) $Y_{est}$ using an ODE solver $\mathbf{SOLVE}(Y(0),\Theta, U+U_{ex})$ and the ground truth measurements of $Y$. The sparsity in the model is introduced by utilizing dropouts in the dense layer.

\noindent{\bf Intuition for addressing C5:} From the dense layer we keep $m$ nodes whose outputs are used to shift the input data $U_{ex}$. Such dense layer nodes can potentially search through a large set of input shifts due to synchronization or reporting errors that best fits the data.  

\noindent{\bf Case study contribution:} We compare LTC-NN-MR with state-of-the-art nonlinear model recovery technique, SINDYc~\cite{kaiser2018sparse}, and other baselines that use the same advanced architecture as LTC-NN-MR but use CT-RNN or NODE as neural nodes. For benchmark case studies, we use three simulation and one real-world examples available in~\cite{kaiser2018sparse}. We introduce two new medical benchmarks with automated insulin delivery (AID) system and electro-encephalogram (EEG) brain signal reconstruction using nonlinear oscillator models~\cite{ghorbanian2015stochastic}. For AID we create simulation benchmark using the Food and Drug Administration (FDA) approved UVA/PADOVA Type 1 Diabetes (T1D) simulator~\cite{visentin2018uva}. We also test LTC-NN-MR on real-world clinical study using the publicly available LOIS-P dataset for pregnant women with T1D~\cite{o2021longitudinal}. For EEG reconstruction case study, we create a simulation benchmark and also test LTC-NN-MR on real-world CHB-MIT Scalp EEG dataset~\cite{guttag2010}. In summary we evaluate our strategy on \textbf{five} simulation and \textbf{three} real-world case studies. 

\vspace{-0.1 in}\section{Related Works}
Table~\ref{tbl:relworks} summarizes the recent works for model recovery. In the linear domain, system identification techniques such as Ho Kalman or Eigen system realization algorithm (ERA)~\cite{oymak2021revisiting} attempt to fit a linear model to data. Such techniques do not scale to nonlinear dynamics, and do not preserve the sparsity structure (violate C3).

The seminal work on extracting nonlinear model from data used stratified symbolic regression and genetic programming~\cite{schmidt2009distilling}. This approach does not scale well with increasing dimensions, and requires measurement of all the state variables (violates C4). 

Significant breakthrough was achieved through introduction of sparse identification of nonlinear dynamics (SINDy) that iteratively selects dominant candidate from a library of high-dimensional nonlinear functions~\cite{quade2018sparse}. The sparsity (constraint C3) was achieved through a sequential threshold ridge regression (STRidge) algorithm that iteratively determines the sparse solution utilizing hard thresholds (manual configuration parameter). The original technique assumes an unperturbed system (violates C2 \& C5), and requires measurements of all the state variables $x_i$ (violates C4). Subsequently SINDy has been extended to SINDYc to tackle constraint C2 for control inputs~\cite{kaiser2018sparse}, however, it still violates C5. 

Attempts have been made to incorporate extraction of implicit models using the SINDy strategy~\cite{kaheman2020sindy}, however, the unmeasured state variables are only limited to the differentials of the original state variables ({\it weakly} implicit). SINDy has been also extended to tackle uncertainty in the magnitude of the state variables~\cite{kaiser2018sparse}, but extension to address timing uncertainties has not been explicitly explored. 

Physics informed Neural Networks (PINN) utilize the concept of automatic differentiation to perform accurate forward and inverse analysis of nonlinear physics models and has been used in many practical domains~\cite{chen2021physics}. However, such models are black-box and cannot provide $\Theta_{est}$ while maintaining the constraints C3 and C4. Recently, PINNs have been integrated with sparse regression to recover model coefficients~\cite{chen2021physics}. Such approaches are not applied for perturbed systems (violates C2, consequently C5), and do not consider the implicit dynamics (C4). A major assumption in these approaches is the knowledge of physics loss for the original model coefficients $\Theta$. This is an impractical (potentially circular) assumption since the original model coefficients are unknown in real-world examples. Recently, NODE structure has been used for forecasting while maintaining metriplectic structures~\cite{Kookjin21Machine}, i.e. algebraic structures in models induced by laws of physics such as energy conservation, first and second law of thermodynamics~\cite{Kookjin21Machine}. Such approaches violate C1, with unrealistically high  sampling rates, violates C2 (and C5) since it uses unperturbed system, and violates C4, no implicit dynamics. 

\noindent{\bf How is this paper different?} We address the model recovery problem in the real world satisfying constraints C1 through C5:

\noindent{\bf Given:} i) A set of sampled trace $\mathcal{T} = \{\mathbf{t}_i\}$, where $ \mathbf{t}_i = \{(Y(0),U(0)) \ldots (Y(k\tau),U(k\tau))\} \bigcup \{(U^1_{ex},t_1),$ $\ldots$ $ (U^q_{ex},t_q)\}$, where $k$ is the number of samples and $\tau$ is the Nyquist sampling rate. 
ii) a ``black box" ODE Solver $Y_{est} = \mathbf{SOLVE}(Y(0),\Theta_{est}, U+U_{ex})$, that takes any $\Theta_{est}$ as input and solves the ODE describing the physics model to provide an estimated / reconstructed $Y_{est}$. 

\noindent{\bf Find:} $\Theta_{est}$ such that $||Y-\mathbf{SOLVE}(\Theta_{est})||^2 \leq \psi$, for some $\psi > 0$

\noindent{\bf Under:} C1 through C5 constraints.


\vspace{-0.1 in}\section{Theoretical foundations}
The forward pass of LTC-NN node is given by~\cite{hasani2021liquid}:
\begin{equation}
\label{eqn:ff}
\scriptsize
    \dot{h}(t) = -h(t)/\rho + f_{NN}(h(t),I(t),t,\omega)(A-h(t)),
\end{equation}
where $h(t)$ is one hidden state of the LTC-NN, $\rho$ is a time constant parameter, required to assist any autonomous system to reach equilibrium state. As such existence of the $-h(t)/\rho$ is an important stability criteria as it ensures that the unperturbed physical system settles in time. $f_{NN}$ is the computation function of each node and is a function of the hidden states, $I(t)$ is the input to the LTC-NN, $\omega$ and $A$ are architecture parameters. 

\begin{remark}
\label{rmk:remark1}
The forward pass of an LTC-NN architecture generates a set of implicit physical dynamics that are equivalent to a bi-linear approximations of the control affine autonomous system in Eqn.\@\ref{eqn:sys}. 
\end{remark}

\noindent{\bf Supporting argument:} Algebraic manipulation of the forward pass of LTC-NN architecture gives the structure of Eqn.\@\ref{eqn:Str} which allows an input dependent time constant $\frac{\rho}{1+\rho f_{NN}(h(t),I(t),t,\omega)}$.

\begin{equation}
    \label{eqn:Str}
    \scriptsize
    \dot{h}(t) = -{h(t)}/\big({\frac{\rho}{1+\rho f_{NN}(h(t),I(t),t,\omega)}}\big) + f_{NN}(h(t),I(t),t,\omega)(A).
\end{equation}

The stability criteria for any autonomous system requires the control affine model to have a time constant term as shown in Eqn. \ref{eqn:tc}
\begin{equation}
    \label{eqn:tc}
    \scriptsize
    \dot{X} = -X/\rho + f_{-\rho}(X) + g(X)U_T,
\end{equation}
where $\rho$ is the time constant of the system and $f_{-\rho}(.)$ is the unperturbed dynamics after removing the time constant component.

Assuming that the autonomous system is a dynamic causal system, the bi-linear approximation~\cite{Friston03Dynamic} of the control affine system in Eqn. \ref{eqn:tc} results in Eqn. \ref{eqn:bilin}.
\begin{equation}
    \label{eqn:bilin}
    \scriptsize
    \dot{X} \approx -X/\rho + f_{-\rho}(X) + B X + CU_T + \sum_j{u^j_TD^jX} + H,
\end{equation}
where $B = \frac{\partial (g(X)U_T)}{\partial X}$, $C = \frac{\partial (g(X)U_T)}{\partial U_T}$, and $D^j = \frac{\partial^2 (g(X)U_T)}{{\partial X}{\partial u^j_T}}$, $H$ is a constant. Rearranging Eqn. \ref{eqn:bilin}, we have the similar form as the LTC-NN forward pass in Eqn. \ref{eqn:SimLTC}.
\begin{equation}
    \label{eqn:SimLTC}
    \scriptsize
    \dot{X} \approx -{X}/\big({\frac{\rho}{1+\rho (B + \sum_j{u^j_T D^j})}}\big) + (f_{-\rho}(X) + CU_T + H).
\end{equation}
We observe that Eqn. \ref{eqn:SimLTC} is the same form as Eqn. \ref{eqn:Str} if the input to the LTC-NN $I(t)$ is a concatenation of $Y$ and $U_T$. The hidden layers of the LTC-NN model an inflated set of implicit dynamics which may include the unmeasured system variables of the physics model. 

\begin{remark}
    \label{rmk:remark2}
    The inflated set of implicit dynamics modeled by LTC-NN induces an over-determined set of equations in the coefficients of the bi-linear approximation of control affine model.
\end{remark}

\noindent{\bf Supporting argument:} The training process of LTC-NN fixes weights and instantiates the hidden layer outputs. The values of the unmeasured variables in $X$ is estimated by the hidden state in each training step utilizing the forward pass and learned LTC-NN weights $\omega$. Hence each forward pass provides an over-determined set of linear equations in the coefficients $B$, $C$, and $D^j$.

The original control affine model coefficients $\Theta$ are nonlinear functions of the coefficients $B$, $C$, and $D^j$s, The dense layer is best suited for exploring a large set of possible nonlinear combinations of $B$, $C$, and $D^j$ that express $\Theta$. An over-determined system of equations is inconsistent and may be unsolvable. The dense layer guided by the ODE solver induced loss function ({\tt ODELoss}) learns a consistent set of linear equations in $B$, $C$,  and $D^j$ and also learns their nonlinear combination to determine $\Theta$.

\vspace{-0.1 in}\subsection{Addressing C1: Sampled Data}
Architectures that enable automatic differentiation such as CT-RNN~\cite{hasani2022closed} or NODE~\cite{chen2018neural} or LTC-NN~\cite{hasani2021liquid} are shown to be capable of simulating arbitrary sampling rates between two time samples of the real data through usage of state-of-art ODE solvers. Hence, such architectures are at a relative advantage over sparse identification mechanisms such as SINDy. In our examples we show that decreasing sampling rate up to the minimum required Nyquist rate~\cite{vaidyanathan2001generalizations} reduces accuracy of model recovery for SINDy class of approaches but has little to no effect on CT-RNN, NODE or LTC.

\vspace{-0.1 in}\subsection{Addressing C2: Perturbed system}
The LTC-NN forward pass has an input dependent time constant. This helps in modeling the perturbed system as demonstrated by the equivalence of Eqn. \ref{eqn:Str} and \ref{eqn:SimLTC}. This is not the case for other neural architecture with automatic differentiation such as  CT-RNN~\cite{hasani2022closed} or NODE~\cite{chen2018neural} (Eqn. \ref{eqn:CTNode}). They do not have a direct mapping to the bi-linear approximation of the control affine system, as is evident by the following equations: 

\begin{scriptsize}
\begin{eqnarray}
\label{eqn:CTNode}
   & \text{CT RNN: } \dot{h}(t) = -\frac{h(t)}{\rho} + f_{CT}(h,I,t,\omega_{CT}),\\\nonumber
   & \text{NODE: } \dot{h}(t) = f_{CT}(h,I,t,\omega_{CT}).
\end{eqnarray}
\end{scriptsize}
\vspace{-0.3 in}\subsection{Addressing C3: Sparsity Preservation}
Sparsity preservation is achieved through the training loss function and dropouts in dense layer. PINNs~\cite{chen2021physics} and NODE~\cite{chen2018neural} achieve sparsity by integrating a ``physics loss". It uses the original model coefficients $\Theta$ in an ODE solver to determine the ground truth $Y$ to compute the RMSE of the estimated $Y_{est}$. While this approach is appropriate in simulation settings, it is not practical since the original model coefficient $\Theta$ is not available in the real world. The presented approach uses ODE loss which uses the estimated model coefficients $\Theta_{est}$ and initial value of measured variables $Y(0)$ to compute the $k - 1$ samples of $Y_{est}$ at times $\{\tau, 2\tau \ldots k\tau\}$. For this purpose, it uses a state-of-art ODE solver that implements the control affine structure of the autonomous system. It computes RMSE with the real data $Y$ as loss. Hence, we do not use $\Theta$ in the training of LTC-NN-MR.

\vspace{-0.1 in}\subsection{Addressing C4: Implicit Dynamics}
Remark \ref{rmk:remark1} shows that LTC-NN forward pass can provide an inflated set of implicit dynamics. This is also true for CT-RNN, and NODE. LTC-NN has the advantage over all other techniques in that it can model implicit dynamics in presence of external inputs.

\vspace{-0.1 in}\subsection{Addressing C5: Input uncertainty}
Input uncertainty is of two types: a) magnitude, and b) temporal. The magnitude uncertainty is inherently handled in neural architectures through the weight update process. To tackle timing uncertainty, a subset $\Delta: |\Delta| = q$ of the dense layer outputs are transformed to the range $[0, 1]$ using a Sigmoid activation function. Each $d_i \in \Delta$ acts a shift operation for the input $u^i_{ex} \in U_{ex}$. Each input $u^i_{ex}$ is shifted by the amount $d_i \times k$ and is used in the input layer as well as the ODE Solver in the loss function for each forward pass of the neural structure. The dense layer is used to search for the set of possibilities of input shifts due to temporal uncertainty.

\vspace{-0.1 in}\section{Implementation}\label{sec:Impl}
\begin{figure}[t]
\center
\includegraphics[trim=0 50 0 0,width=\columnwidth]{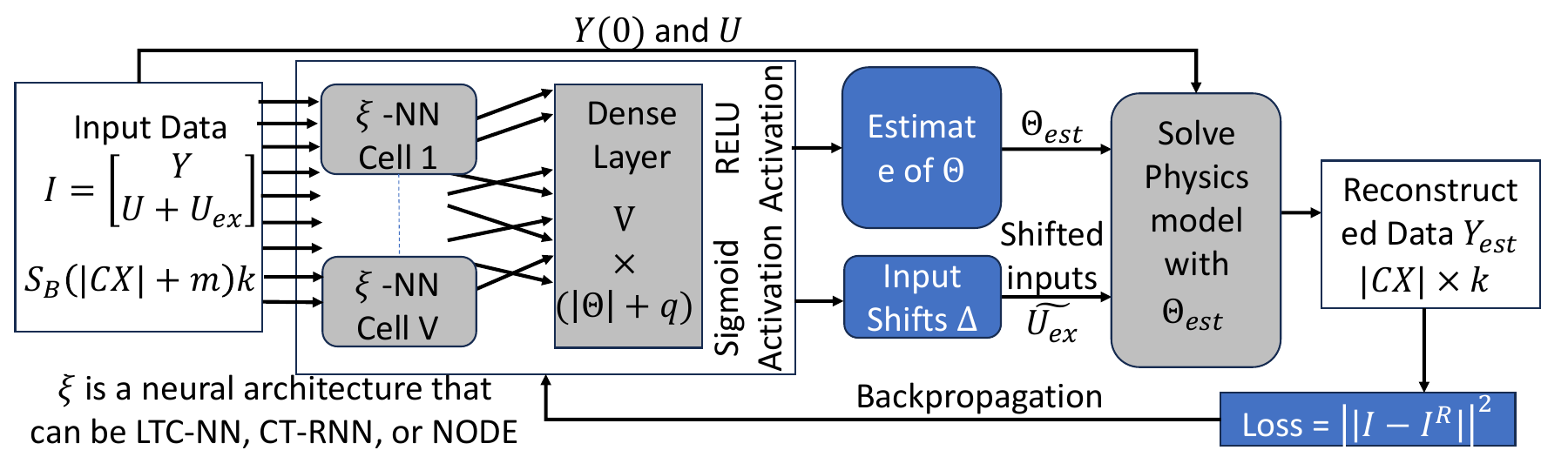} 
 \caption{Implementation of neural architecture based model recovery solution with input uncertainty.}
    \label{fig:Impl}
\end{figure} 
The advanced neural architectures for model recovery ($\xi$-MR, where $\xi$ is either LTC-NN, CT-RNN, or NODE) (Fig. \ref{fig:Impl}) is implemented by extending the base code available in~\cite{liquid-time-constant-networks}. For each example, we extract the training data consisting of temporal traces of $Y$, $U$ and $Ex$. $Y$ is sampled at least at the Nyquist rate for the application, and $U$ has the same sampling rate as $Y$. $Ex$ is a set of tuples denoting $U^i_{ex}$ values at time $t^i_{ex}$. $Ex$ is transformed into a signal at the same sampling frequency as $Y$ by keeping $U^i_{ex}$ at times $t^i_{ex}$ and appending 0s at all other times. The resulting training data is then divided into batches of size $S_B$ forming a 3 D tensor of size $S_B \times (|Y|+m) \times k$. 

\begin{table*}
	\centering
	\scriptsize
	\caption{Benchmark Examples. (B) denotes examples available in ~\cite{kaiser2018sparse}, (N) denotes novel case studies introduced in this paper.}
	\begin{tabular}{p{1.8 in}|p{0.75 in}|p{0.2 in}|p{0.3 in}|p{0.7 in}|p{0.5 in}|p{0.75 in}|p{0.7 in}}
	 \hline
		\textit{Example} & \textit{Variables} & \textit{Inputs} &  \textit{Implicit}  & \textit{Uncertain timing} & \textit{Nyquist rate} & \textit{Max sampling rate} & \textit{No of coefficients}  \\ \hline
Real-World: Lotka Volterra (B) & $x_1$, $x_2$ & 1 &	$x_1$ & No & 2.5 Hz & 10 Hz & 4\\
Simulation: Chaotic Lorenz System (B) & $x_1$, $x_2$, $x_3$ & 1 & $x_1$ $x_2$ &	No & 100Hz & 1000 Hz & 4\\
Simulation: F8 Crusader tracking (B)& $x_1$, $x_2$, $x_3$  & 1  &  $x_2$   & No & 100 Hz  & 1000 Hz & 20\\
Simulation: Pathogenic attack (B)& $x_1$, $x_2$, $x_3$ $x_4$, $x_5$ & 1 & $x_3$, $x_4$ &	No & 2.8 $\times$ $10^{-4}$ & 5.6 $\times$ $10^{-4}$ & 13\\
Real-world: Automated Insulin Delivery (N) & $I_s$ $G$ $I$ & 2 & $I_s$, $I$ &	Yes & 0.0028 Hz & 0.0033 Hz & 9\\
Simulation: AID (N) & $I_s$ $G$ $I$ & 2 & $I_s$, $I$ &	Yes & 0.0028 Hz & 10 Hz & 9\\
Real -world: EEG (N) & $x_1$,$x_2$,$\dot{x}_1$,$\dot{x}_2$ & 1 & $x_1$, $\dot{x_1}$ & Yes & 250 Hz & 500 Hz & 6\\
		\bottomrule
	\end{tabular}
	\label{tbl:examples}
\end{table*}

Each batch is passed through the $\xi$ network with $V$ nodes, resulting in $V$ hidden states. A dense layer is then employed to transform this $V$ hidden states into $p = |\Theta|$ model coefficient estimates and $q$ input shift values. The dense layer is a multi-layer perceptron with ReLU activation function for the model coefficient estimate nodes and Sigmoid activation function for input shift values. The input shift values are used to shift the external input vector. The shifted inputs, the model coefficient estimates, and the initial value $Y(0)$ is passed through an ODE solver, that solves the control affine model in Eqn. \ref{eqn:sys} with the coefficients $\Theta_{est}$, initial conditions $Y(0)$ and inputs $U$ and $U_{ex}$. The Runge Kutta integration method is used in the ODE solver, which gives $Y_{est}$. The backpropagation of the network is performed using the network loss appended with ODE loss, which is the mean square error between the original trace $Y$ and estimated trace $Y_{est}$.

\vspace{-0.1 in}\section{Evaluation}
Experiments are performed in Nvidia Titan V GPU, CUDA 11.4 and TensorFlow 2.7.0~\cite{tensorflow-releases}. Code and data available in~\cite{supp}. We show two types of examples: 

\noindent a) \textbf{Five} simulation benchmarks, three obtained from SINDYc~\cite{kaiser2018sparse} and the AID and EEG problems introduced in this paper. Simulation is used to test the variation of model recovery performance under various severity levels of the constraints.

\noindent b) \textbf{Three }real-world data available for the example of Lotka Volterra system in the original SINDYc paper~\cite{kaiser2018sparse}, the LOIS-P data for AID in pregnancy~\cite{o2021longitudinal}, and the EEG data for epilepsy~\cite{guttag2010}. This shows practical applicability of LTC-NN-MR in real-world data. 
\vspace{-0.1 in}\subsection{Benchmark Examples}
 Table \ref{tbl:examples} shows all examples (physical dynamics are in supplementary document~\cite{supp}). In the examples from~\cite{kaiser2018sparse} marked with (B) in Table \ref{tbl:examples}, there are no external human inputs $U_{ex}$ apart from control inputs $u$. We arbitrarily varied the input timing to evaluate effect of input uncertainty constraint C5. For the AID example in simulation, we alter the meal timing. For the EEG example in simulation we chose a deterministic sinusoidal input and then a stochastic Wiener process generated input to test for high levels of input uncertainties. In the real world EEG and AID example uncertainty is inherent in the data. 
\vspace{-0.2 in}\subsubsection{Automated Insulin Delivery System}
In the AID system, the glucose insulin dynamics is given by the Bergman Minimal Model (BMM)~\cite{bergman2021origins}:
\begin{scriptsize}
\begin{eqnarray}
\label{eqn:5}
& &\dot{i}(t) = -n i(t) + p_4 u_1(t), \dot{i}_s(t) = -p_1 i_s(t) + p_2 ( i(t) - i_b)\\
\label{eqn:7}
& & \dot{G}(t) = - i_s (t) G_b -p3 ( G(t)) + u2(t)/VoI,
\end{eqnarray}
\end{scriptsize}
The input vector $U(t)$ consists of the input insulin level $u_1(t)$, which is derived using a self adaptive MPC controller like Tandem Control IQ~\cite{Lamrani21TII}. $U_{ex}$ consists of the glucose appearance in the body $u_2$ for a meal. In the real world, users forget to report the exact timing of the meal and also make mistake in estimating consumed carbohydrate amount~\cite{Lamrani21TII}. We model this error as input uncertainty in time and magnitude. The state vector $X(t)$ comprises of the blood insulin level $i$, the interstitial insulin level $i_s$, and the BG level $G$. The measured vector $Y = G$ since CGM measures only glucose. $p_1$, $p_2$, $i_b$, $p_3$, $p_4$, $n$, and $1/V_o I$ are all patient specific coefficients.
\vspace{-0.1 in}
\subsubsection{EEG example}
The EEG represents brain waves and can be modeled using coupled system of Duffing—van der Pol oscillators~\cite{ghorbanian2015stochastic} as in Eqn. \ref{eqn:EEG}.

\begin{scriptsize}
\begin{eqnarray}
\label{eqn:EEG}
    \ddot{x}_1 + k_1x_1  =k_2x_2 -b_1x_3^2 - b_2(x_1 - x_2)^3 + \epsilon_1 \dot{x}_1 (1-x_1^2) \\\nonumber
\ddot{x}_2 - k_2(x_1 - x_2) = b_2(x_1 - x_2)^3 + \epsilon_2 \dot{x}_2(1-x_2^2) + \mu dW,
\end{eqnarray}
\end{scriptsize}
\noindent \!\!\!\! where $k_1$, $k_2$, $b_1$,$b_2$, $\epsilon_1$ and $\epsilon_2$ are patient specific parameters, $x_2$ is the EEG signal, $x_1$ is an internal unmeasured state variable, $\mu$ is the magnitude of activation, $dW$ is a random variable following the Wiener process~\cite{ghorbanian2015stochastic}. Here the input $u =\mu dW$ is a stochastic process with significant temporal and magnitude uncertainty.

\vspace{-0.1 in}\subsubsection{Real-World Data}
Three examples of real-world data are shown in this paper:

A) Lotka Voltera system uses yearly lynx and hare pelts data collected from Hudson Bay Company~\cite{kaiser2018sparse}. 

B) In the real-world AID example we used the LOIS-P dataset~\cite{o2021longitudinal}, which is data from a clinical study on 25 patients with pre-existing T1D for at least a year. All patients were enrolled before 17 weeks gestational age at three sites: Mayo Clinic, Rochester, Mount Sinai in New York City, and Sansum Diabetes Research Institute. On an average 24.7 weeks ($\pm$ 5.2) of Dexcom G6 CGM glucose at 5 min interval and insulin and meal intake data. 

C) The EEG example is solely evaluated in the real world with the CHB-MIT Scalp EEG database~\cite{guttag2010}. It has 684 EEG signals from 22 pediatric subjects with epilepsy with a sampling rate of 250 Hz.

\vspace{-0.1 in}\subsubsection{Simulation data}
Benchmark simulations use data from of SINDYc available at~\cite{sindy-mpc}.

\noindent{\bf Simulation data for AID:} We considered 14 traces of glucose insulin dynamics each of which were 200 samples (16 hrs). In each trace meal ingestion time was varied from [t = 15 min to t = 400 min] with carbohydrate value randomly sampled from the range $[0g, 28g]$ for each meal, and bolus insulin delivery was sampled from the set $[0 U, 40 U]$. The traces were generated using the T1D simulator~\cite{visentin2018uva}. Two sampling frequencies are tested: i) 10Hz, and ii) real-world CGM sampling rate of every 5 min (Table \ref{tbl:examples}). 

\vspace{-0.1 in}\subsection{Baseline Techniques}
We compare with the following baseline strategies:

\noindent{\bf SINDYc:} This baseline~\cite{kaiser2018sparse} is used to show that reducing sampling rate to a minimum of Nyquist rate, causes significant degradation of performance for SINDY class of approaches. 

\noindent{\bf NODE-MR:} This baseline is the seminal work on neural architecture~\cite{Kookjin21Machine}. It is used to show that lack of a time constant factor in the forward pass (Eqn. \ref{eqn:CTNode}) reduces accuracy in recovering model. 

\noindent{\bf CT-RNN-MR:} This baseline strategy has an input independent time constant factor (Eqn. \ref{eqn:CTNode}). This is used to show that although time constant factor independent of input in the forward pass cannot accurately recover model from a perturbed system.


\vspace{-0.1 in}\subsection{Evaluation experiments and metrics of success}
For each evaluation experiment, we use two metrics:

\noindent{\bf Root mean square error in model coefficients ($RMSE_\Theta$):} Given the estimated model coefficients $\Theta_{est}$ we compute:
\begin{equation}
    \label{eqn:rmse}
    \scriptsize
    RMSE_\Theta = \sqrt{1/p \times \sum_{j = 1 \ldots p}{(\Theta^j_{est}-\Theta^j)^2}}.
\end{equation}
\noindent{\bf Root mean square error in signal ($RMSE_Y$):} Given the estimation of the measured variables $Y_{est}$ for any technique we compute:
\begin{equation}
    \label{eqn:rmsey}
    \scriptsize
    RMSE_Y = 1/n\sum_{l = 1 \ldots n}{\sqrt{1/k \times \sum_{j = 1 \ldots k}{(Y^l_{est}(j)-Y^l(j))^2}}}.
\end{equation}
\begin{figure*}
\center
\includegraphics[clip=false,trim=0 25 0 0,width=0.75\textwidth]{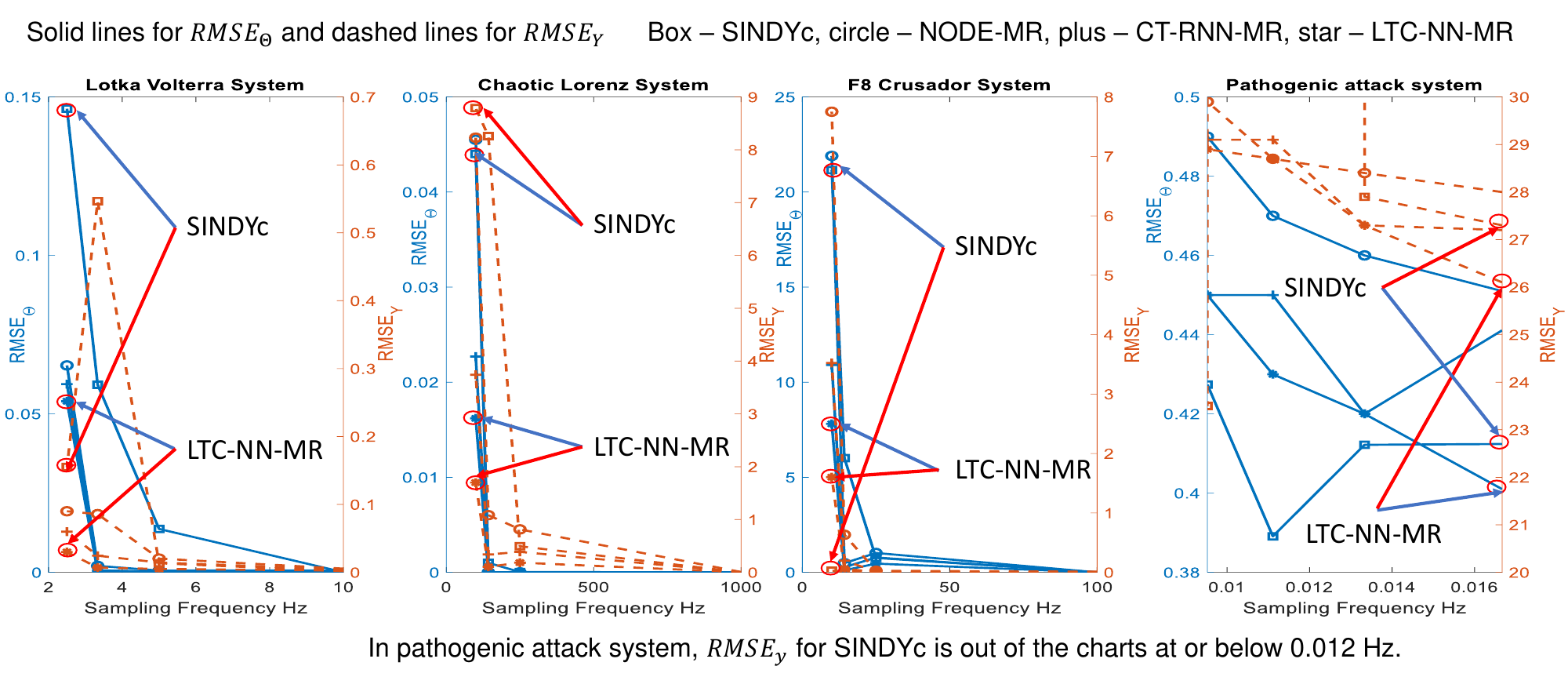} 
 \caption{Comparison of SINDYc with neural architecture for configuration $\Phi_N$ by only varying sampling frequency to Nyquist rate, no implicit dynamics, with perturbation, and no input uncertainty.}
    \label{fig:C1}
\end{figure*} 
All compared techniques identify sparsity preserving dynamics. We start with a configuration $\Phi_0$ that has high sampling rates shown in column 7 of Table \ref{tbl:examples}, has no implicit dynamics, has input perturbation, only uncertainty in input magnitude (no temporal uncertainty). We conduct the following evaluation experiments:

\noindent{\it Effect of sampling rate (C1) on (B) examples:} We vary the sampling rate of $\Phi_0$ from the rate used in the simulation data to the Nyquist rate (Table \ref{tbl:examples}) and analyze the variation of $RMSE_\Theta$ and $RMSE_Y$. The configuration with Nyquist sampling rate is denoted as $\Phi_N$. 

\noindent{\it Effect of perturbation (C2) on (B) examples:} From $\Phi_N$, we create configuration $\Phi_{NI}$ by removing the input perturbation from model and data. Effect of perturbation is the difference in performance of $\Phi_N$ and $\Phi_{NI}$.

\noindent{\it Effect of implicit dynamics (C3) on (B) examples:} From $\Phi_{N}$ we created another configuration $\Phi_{NP}$ where measurements of implicit dynamics are withheld. This experiment compares neural architectures based on their capability in searching for implicit dynamics. We kept input perturbation. In this comparison, we do not include SINDYc since it is not designed to extract implicit dynamics.

\noindent{\it Effect of sparsity (C4):} The techniques that we evaluate all maintain sparsity. Hence, C4 is evaluated in conjunction with C3.

\noindent{\it Effect of input uncertainty (C5) on (B) and (N) simulation examples:} For each case study, we vary the time stamp of input $u$ from 3 samples to 20 samples while keeping the reported time stamp the same.

\noindent{\it Real-world experiments on AID (N) and EEG (N):} We compare the three neural architecture for their performance in modeling real data. For this experiment, we do not know the ground truth model coefficients $\Theta$ and we compare the techniques using $RMSE_Y$.

\vspace{-0.1 in}\subsection{Training and Validation Method}

\noindent{\bf SINDYc:} We utilize the same training method as used in SINDYc and as reflected in the code accessed from~\cite{sindy-mpc}. In the experiments, with $\phi_N$, we changed the \texttt{dt} variable in the ``getTrainingData.m" from the minimum value corresponding to maximum frequency in each example (Table \ref{tbl:examples}) to the Nyquist rate. The Nyquist rate is obtained by computing the power spectral density of the signals sampled at the highest frequency. We then extract the frequency $f_{90}$ at which the cumulative power density reaches 90\% of the maximum level. The Nyquist rate is two times $f_{90}$. We change the $dt$ to obtain four frequency points up to the maximum $dt$ for the Nyquist rate. For each example, the training and validation implemented in~\cite{sindy-mpc} used to generate the $RMSE_Y$ and $RMSE_\Theta$ for SINDYc.

\noindent{\bf Neural Architectures:} For the neural architectures, we based our implementation on the codebase available at~\cite{liquid-time-constant-networks}. Here a generic framework for LTC-NN, CT-RNN, and NODE is implemented using TensorFlow 2.7.0. We wrote a custom loss function (code available in ~\cite{supp}) that implements the Runge Kutta solution of the physical dynamics given a vector of model coefficients. We use the general training architecture presented in~\cite{liquid-time-constant-networks} with an ADAM optimizer. The framework can be instantiated with LTC-NN, CT-RNN and NODE core architecture through an input parameter. Batch training was utilized for each example. For each example, we took the same simulation data as SINDYc and divided the traces into 48 instance of training and 16 instance of test each of at least $k=200$ samples. These training instances were passed to the neural architectures with a batch size $S_B=32$. The $RMSE_Y$ and $RMSE_\Theta$ are reported on the test data. 

\vspace{-0.1 in}\subsection{Results}
We first compare SINDYc and all neural architecture on the benchmark examples in~\cite{kaiser2018sparse}. We evaluate C5 with the simulation AID example. Since SINDYc does not model implicit dynamics, we only compare the neural architecture for real-world AID example.

\vspace{-0.1 in}\subsubsection{Benchmark examples}

\noindent{\bf Effect of sampling rate (C1):} As seen from Fig. \ref{fig:C1}, at sampling rates nearly four times Nyquist rate, all techniques give similar $RMSE_\Theta$ and $RMSE_Y$. As sampling rates are increased every technique has degradation in both the performance metrics. However, SINDYc is most affected by the change in sampling frequency. All neural architectures perform better than SINDYc with LTC-NN-MR showing the best performance. The primary reason for such a result is the fact that as data is sampled less frequently, the set of potential models that fit the data increases. While SINDYc imposes the constraint of sparsity, the neural architectures impose further constraints on top of sparsity through their structure. The most restrictive constraint is the input dependent time constant of LTC-NN-MR and hence it performs the best. CT-RNN-MR has the next most stringent constraint while NODE has the least restrictive constraint among the neural architectures. In all examples except for the pathogenic attack, we see similar trends for $RMSE_\Theta$, where LTC-NN-MR outperforms all neural architectures, which in turn outperforms SINDYc at Nyquist sampling rate. For the pathogenic attack example, an interesting occurrence is observed, where at Nyquist rate, SINDYc has the best performance in both the metrics. However, at the next sampling frequency we see SINDYc has a very high $RMSE_y$ for a slight change in $RMSE_\Theta$. All neural architectures differed from this trend and both $RMSE_\Theta$ and $RMSE_Y$ improved. The main reason for this is that SINDYc violates the sparsity constraint (C2). On closer look we found that SINDYc found a totally different physical model of the system. A hint of this behavior is also seen in the LOTKA-Volterra and F8 Crusader example, where decreasing sampling frequency to Nyquist rate reduced $RMSE_y$ but increased $RMSE_\Theta$. Similarly it was observed that SINDYc compensated for loss in $RMSE_\Theta$ performance by adding extra nonlinear terms to reduce $RMSE_y$. From the results, we do not see such behavior for the neural architectures. One main reason can be because ODE Solver at the loss function guides the exploration of the dynamics. 

\begin{figure}
\center
\includegraphics[trim=0 0 0 0,width=\linewidth,trim=0 50 0 0]{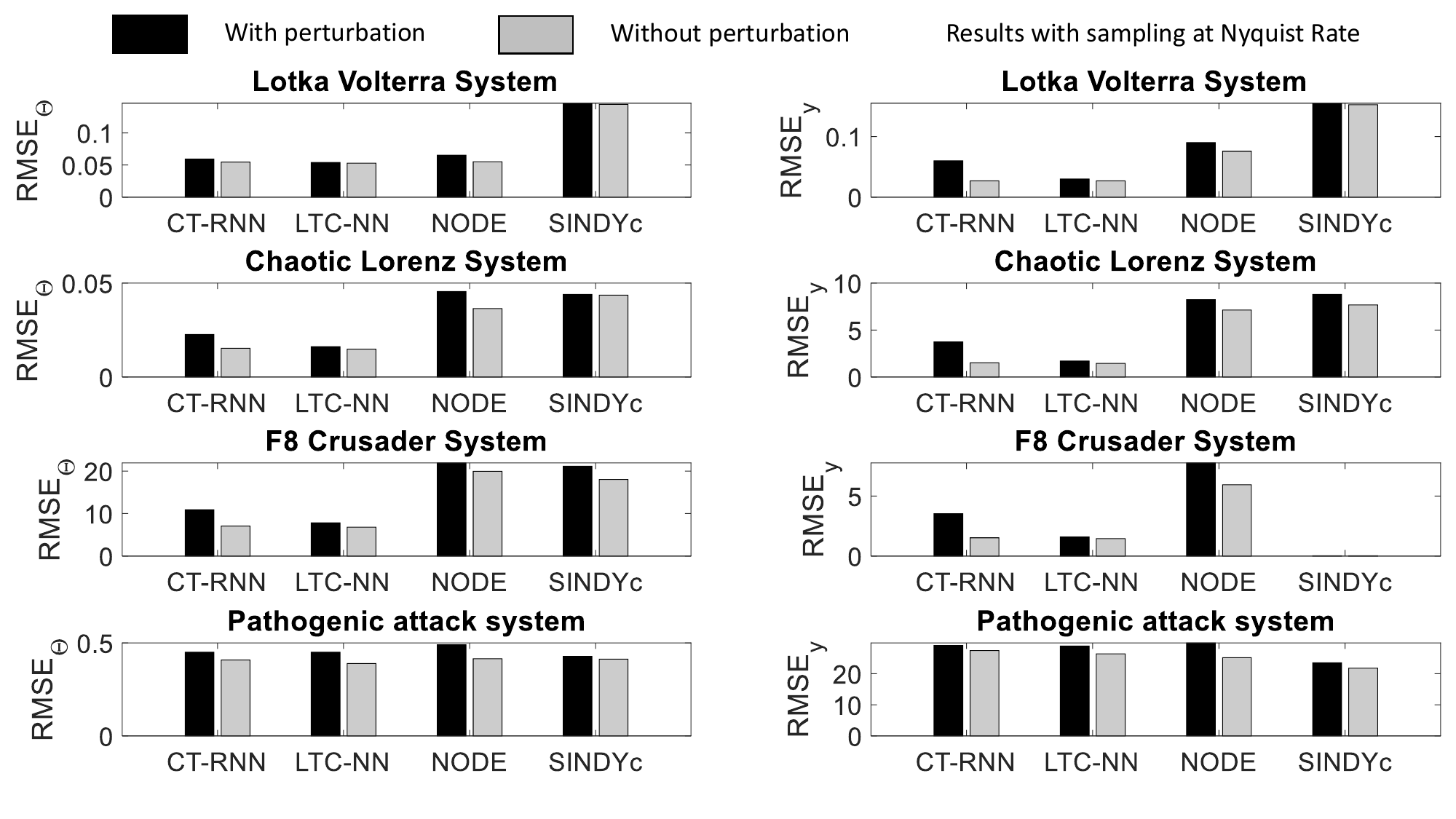} 
 \caption{Comparison of SINDYc with neural architecture for configuration $\Phi_{NI}$ by sampling at Nyquist rate, no implicit dynamics, with/ without perturbation, and no input uncertainty.}
    \label{fig:C2}
\end{figure}

\noindent{\bf Effect of perturbation (C2):} Fig. \ref{fig:C2} shows that when input perturbation is removed, all techniques have improved performance in both metrics. Interestingly, we see that CT-RNN-MR without input perturbation has similar performance to LTC-NN-MR architecture. This agrees with the theory since, without input the forward pass of CT-RNN-MR and LTC-NN-MR are similar. On the other hand, although NODE-MR had improvement in performance but it could not match CT-RNN-MR and LTC-NN-MR performance for no input perturbation. SINDYc also had performance improvement when input perturbation was removed. 

\noindent{\bf Effect of implicit dynamics (C3 + C4):} We updated the loss function of each architecture as follows:
\begin{equation}
    \label{eqn:LossUpdate}
    \scriptsize
    Loss = 1/n\sum_{i = 1 \ldots n}{\sqrt{1/k\sum_{j = 1 \ldots k}{|x_i(k) - x^{est}_i(k)|^2}}},
\end{equation}
where $x^{est}_i \in X_{est} = \mathbf{SOLVE}(X(0),\Theta_{est})$. Table \ref{tbl:C3} shows that although there is some improvement in both the parameters, it was expected that $RMSE_\Theta$ should significantly improve by providing measurements of implicit dynamics. However, we do not see such improvements. A possible explanation for this is that since all the baseline examples are observable systems, the implicit dynamics could be derived in terms of the measured system variables. The conjecture is that the neural architectures are capable of modeling the implicit dynamics in terms of the measured variables. This should be further investigated in future works.

\begin{table}
	\centering
	\scriptsize
	\caption{Effect of providing the measurement of implicit dynamics to the neural architectures on benchmark examples.}
	\begin{tabular}{p{0.35 in}p{0.35 in}p{0.225 in}p{0.225 in}p{0.225 in}p{0.225 in}p{0.225 in}p{0.225 in}}
	 \hline
		{Example} & {RMSE} & \multicolumn{2}{c}{LTC-NN-MR} & \multicolumn{2}{c}{CT-RNN-MR} & \multicolumn{2}{c}{NODE-MR}  \\ \hline
   & & {Implicit} & {Explicit}& {Implicit} & {Explicit}& {Implicit} & {Explicit}\\ \hline
Lotka  & $RMSE_\Theta$ & 0.054 & 0.048 & 0.06 & 0.054 & 0.065 & 0.064 \\
Volterra & $RMSE_Y$ & 0.03 & 0.03 & 0.06 & 0.05 & 0.09 & 0.088 \\\hline
Chaotic  & $RMSE_\Theta$ & 0.016 & 0.015 & 0.023 & 0.022 & 0.045 & 0.044 \\
Lorenz & $RMSE_Y$ & 1.7 & 1.68 & 3.74 & 3.66 & 8.23 & 8.1 \\\hline
 F8  & $RMSE_\Theta$ & 7.81 & 6.8 & 10.9 & 10.5 & 21.9 & 19.9 \\
Crusader& $RMSE_Y$ & 1.6 & 1.57 & 3.52 & 3.46 & 7.75 & 7.22 \\\hline
 Pathogenics  & $RMSE_\Theta$ & 0.45 & 0.39 & 0.45 & 0.43 & 0.49 & 0.42 \\
attack & $RMSE_Y$ & 28.9 & 28.3 & 29.1 & 28.8 & 29.9 & 29.5 \\
		\bottomrule
	\end{tabular}
	\label{tbl:C3}
\end{table}

\noindent{\bf Effect of violating input uncertainty (C5):} Each input in the benchmark examples were shifted by 2 to 20 samples. However, the model recovery methods were not aware of this change and assumed that the inputs occurred at the designated time. Table \ref{tab:R2} shows the average degradation in $RMSE_y$ and $RMSE_\theta$ when the input shifts are disabled for LTC-NN-MR. The degradation is significantly reduced when input shifts are re-introduced. 

\begin{table}
	\centering
	\scriptsize
	\caption{Percentage degradation of $RMSE_\theta$, $RMSE_y$, for C5 violation and recovery with input shifts. SC: SINDY-c, LM: LTC-NN-MR, LV: Lotka Volterra, CL: Lorenz, F8: Cruiser: PA: Pathogenics.}
	\begin{tabular}{p{0.1 in}p{0.15 in}p{0.15 in}p{0.15 in}p{0.15 in}p{0.725 in}p{0.725 in}}
	 \hline
		{Case} & \multicolumn{2}{c}{$RMSE_y$} & \multicolumn{2}{c}{$RMSE_\theta$} & $RMSE_y$ + shifts & $RMSE_\theta$ + shifts   \\ 
   & {SC} & {LM} & {SC} & {LM }& {LM} & {LM }\\ 
LV  & 120\% & 56\% & 65\% & 53\% & 8\% & 7\% \\
CL  & 210\% & 121\% & 32\% & 27\% & 13\% & 3\% \\
 F8  & 431\% & 212\% & 111\% & 89\% & 13\% & 7\% \\
 PA  & 78\% & 36\% & 24\% & 22\% & 11\% & 6\% \\ \bottomrule
	\end{tabular}
	\label{tab:R2}
 \vspace{-0.1 in}
\end{table}

\vspace{-0.1 in}\subsubsection{Automated Insulin Delivery Example}
\noindent{\bf Simulation Examples:} In Table \ref{tbl:C4}, SINDYc did not have any temporal uncertainty for meal inputs. The neural architectures had temporal uncertainty at meal inputs and also used input shifts in the architecture. However, for the last column, the input shift from neural architectures are removed and SINDYc is also evaluated for uncertainty at meal input. All techniques perform well for model recovery from simulation data at 10 Hz sampling rate. SINDYc shows excellent $RMSE_Y$ but poor $RMSE_\Theta$. The neural architectures perform similar to SINDYc at such high sampling frequency. However, when the sampling rate is reduced to Nyquist rate, all methods have performance degradation, with SINDYc suffering the most. LTC-NN-MR still performs better than the baseline techniques. When input shifts are removed, all techniques suffered significant performance drop. 

\begin{table}
	\centering
	\scriptsize
	\caption{Comparison of baseline techniques for AID simulation example with no implicit dynamics. It also shows effect of input uncertainty (C5) in last two rows. $f_s$ is sampling frequency}
	\begin{tabular}{p{0.55 in}p{0.25 in}p{0.3 in}p{0.25 in}p{0.3 in}p{0.25 in}p{0.3 in}}
	 \hline
		{Approach} & \multicolumn{2}{c}{$f_s = 10 Hz$} & \multicolumn{2}{c}{$f_s = 0.0033 Hz$} & \multicolumn{2}{c}{without input shifts} \\ \midrule
   & {$RMSE_Y$} & {$RMSE_\Theta$}  & {$RMSE_Y$} & {$RMSE_\Theta$} & {$RMSE_Y$} & {$RMSE_\Theta$} \\
SINDYc  & 0.004 & 0.342 & 14.5 & 2.44 & 101.6 & 22.3  \\
LTC-NN-MR & 0.003 & 0.213 & 0.31 & 0.45 & 31.3 & 14.1  \\
CT-RNN-MR  & 0.007 & 0.311 & 0.76 & 0.8 & 43.2 & 17.8 \\
NODE-MR & 0.012 & 0.56 & 1.3 & 1.1 & 77.4 & 23.6\\
		\bottomrule
	\end{tabular}
	\label{tbl:C4}
\end{table}

\noindent{\bf Real-World Example Input Uncertainty (C5):} Table \ref{tbl:Real} shows the performance of the neural architectures on real data. Without exploring temporal uncertainty of inputs, the best $RMSE_Y$ obtained for LTC-NN-MR was 26.1. Note that the state of art CGM prediction mechanism for 30 mins ahead prediction has an RMSE of 11.1~\cite{deng2021deep}. With input shift enabled in the architecture, we see significant improvement in $RMSE_Y$ for each neural architectures. For LTC-NN-MR we see an $RMSE_Y$ of 3.03 which is significantly better than state-of-art forecasting mechanisms.   

\begin{table}
	\centering
	\scriptsize
	\caption{$RMSE_Y$ comparison for AID real-world example with sampled data (C1), control + human perturbed system (C2), sparse dynamics (C3), implicit dynamics (C4), and input uncertainty (C5)  .}
		\begin{tabular}{p{0.75 in}p{0.5 in}p{0.5 in}}
	 \hline
		{Approach} &  {not C5} & {C5}\\ \hline
NODE-MR  & 45.6 & 8.7 \\
CT-RNN-MR &  32.3 & 6.8\\
LTC-NN-MR  &  26.1 & 3.03 \\
		\bottomrule
	\end{tabular}
	\label{tbl:Real}
\end{table}

\begin{table}
	\centering
	\scriptsize
	\caption{$RMSE_\Theta$ ($R_\Theta$) and $RMSE_y$ ($R_Y$) for EEG simulation}
	\begin{tabular}{p{0.55 in}p{0.25 in}p{0.3 in}p{0.25 in}p{0.3 in}p{0.25 in}p{0.3 in}}
	 \hline
		{Approach} & \multicolumn{2}{c}{Sine input} & \multicolumn{2}{c}{Sine input } & \multicolumn{2}{c}{Wiener input} \\ 
  & \multicolumn{2}{c}{no implicit} & \multicolumn{2}{c}{$x_1$ implicit} & & \\\midrule
   & {$R_Y$} & {$R_\Theta$}  & {$R_Y$} & {$R_\Theta$} & {$R_Y$} & {$R_\Theta$} \\ 
SINDYc  & 0.1 & 0.21 & 23.2 & 46.1 & 144.1 & 101.3  \\
LTC-NN-MR & 0.1 & 0.203 & 6.3 & 4.7 & 19.8 & 12.9  \\
		\bottomrule
	\end{tabular}
	\label{tbl:EEG}
\end{table}

\begin{table}
	\centering
	\scriptsize
	\caption{$RMSE_y$ for EEG real-world example}
	\begin{tabular}{p{0.75 in}p{0.55 in}}
	 \hline
		{SINDYc} & {LTC-NN-MR} \\\midrule
   
1211.3 ($\pm$ 489.1) & 41.2 ($\pm$ 27.9)\\
		\bottomrule
	\end{tabular}
	\label{tbl:EEGreal}
\end{table}

\vspace{-0.1 in}\subsubsection{EEG reconstruction example}
\noindent{\bf Simulation Example:} In simulation, we use a sinusoidal input as activation instead of the Wiener process in~\cite{ghorbanian2015stochastic}. Table \ref{tbl:EEG} shows that both SINDYc and LTC-NN-MR have comparable accuracy in extracting the model coefficients when measurement of both $x_1$ and $x_2$ are available. However, if $x_1$ measurements are withheld, we see that SINDYc recovers an entirely wrong model with high $RMSE_\Theta$, whereas LTC-NN-MR has much lower $RMSE_\Theta$ than SINDYc. Interestingly, if we use the Wiener process as input, then even if we make $x_1$ available to SINDYc it still recovers a wrong model, which is not the case for LTC-NN-MR.

\noindent{\bf Real-World Example:} Table \ref{tbl:EEGreal} shows that LTC-NN-MR can replicate the EEG signal with much lower $RMSE_y$ than SINDYc from all patients in the CHB-MIT Scalp EEG dataset.

\vspace{-0.1 in}\subsubsection{Execution Time}
For the LTC-NN-MR architecture, the computation complexity of forward pass is $O(V+V\times (|\Theta|+q)) + O(|X|N)$, where $N$ is the number of samples in the data, $V$,$q$,$\Theta$,$X$ are described in Section \ref{sec:Impl}. The complexity of backward pass is $O(V\times P_{LTC}\times N + V \times (|\Theta|+q) \times P_{dense} \times N)$, where $P_{LTC}$ is the number of parameters in the LTC cell, and $P_{dense}$ is the number of parameters in each neuron of the dense layer. SINDYc ran on a single CPU thread and was 11.3 ($\pm$ 2.1) times faster than the neural architecture on GPU. However, at in real-world data, SINDYc has much poorer performance than LTC-NN-MR. This accuracy time trade-off has to be carefully explored for a given application. Usage of cloud computing may improve the speed for neural architectures and make them a viable candidate for real time accurate model recovery.

\vspace{-0.1 in}\section{Conclusions}
This paper provided a liquid time constant neural network based solution to the problem of recovering coefficients of a physics model from real-world data from a dynamical system. It identified five key challenges in real-world deployments and provided solutions to mitigate them effectively. The practical challenge of information loss due to low sampling rates (C1) was overcome utilizing the automatic differentiation property of LTC-NN, where each node can reproduce samples with arbitrary precision in between sampling time stamps. LTC-NN-MR handles transients introduced due to discontinuous inputs (C2) by decoupling the input effects from the unperturbed dynamical system through its control affine forward pass dynamics. To derive the sparsest solution (C3), LTC-NN-MR combines ODE solver based reconstruction loss with dense layer dropout and hence does not need specific knowledge about the underlying equations such as sparsity level as in SINDYc or ground truth model coefficients as in PINNs. To the best of our knowledge, LTC-NN-MR is the only technique to recover model coefficients in presence of implicit dynamics (C4) unlike the weak implicit notions in SINDYc extensions. Further, LTC-NN-MR is the only technique to be able to recover an accurate model in the presence of input timing uncertainties making it amenable to be used in safety-critical real-world human-in-the-loop dynamical systems. 

\noindent{\bf Ethical Considerations:} LTC-NN-MR solves an important problem of model recovery in human-in-the-loop and human-in-the-plant systems~\cite{banerjee24HILHIP}. In this domain, one of the applications of LTC-NN-MR is digital twins. An unethical usage is impersonation. Thus, careful ethical evaluation is required when integrating such systems in safety-critical applications e.g. medical practice~\cite{banerjee24Ethics}. 
\bibliography{m2738}

\end{document}